\documentclass[pdflatex,sn-mathphys-num]{sn-jnl}

\usepackage{graphicx}%
\usepackage{multirow}%
\usepackage{amsmath,amssymb,amsfonts}%
\usepackage{amsthm}%
\usepackage{mathrsfs}%
\usepackage[title]{appendix}%
\usepackage{xcolor}%
\usepackage{textcomp}%
\usepackage{manyfoot}%
\usepackage{booktabs}%
\usepackage{algorithm}%
\usepackage{algorithmicx}%
\usepackage{algpseudocode}%
\usepackage{listings}%

\theoremstyle{thmstyleone}%
\theoremstyle{thmstyletwo}%

\theoremstyle{thmstylethree}%

\begin{document}

\title[Article Title]{SCOPE-PD: Explainable AI on Subjective and Clinical Objective Measurements of Parkinson's Disease for Precision Decision-Making}


\author[1]{\fnm{Md Mezbahul} \sur{Islam}}\email{misla093@fiu.edu}

\author[2]{\fnm{John Michael} \sur{Templeton}}\email{jtemplet@usf.edu}

\author[1]{\fnm{Masrur} \sur{Sobhan}}\email{msobh002@fiu.edu}

\author[1]{\fnm{Christian} \sur{Poellabauer}}\email{cpoellab@fiu.edu}

\author*[1]{\fnm{Ananda Mohan} \sur{Mondal}}\email{amondal@fiu.edu}

\affil*[1]{\orgdiv{Computing \& Information Sciences}, \orgname{Florida International University}, \orgaddress{\street{11200 SW 8th St}, \city{Miami}, \postcode{33199}, \state{Florida}, \country{USA}}}

\affil[2]{\orgdiv{Computer Science \& Engineering}, \orgname{University of South Florida}, \orgaddress{\street{3820 USF Alumni Drive}, \city{Tampa}, \postcode{33620}, \state{Florida}, \country{USA}}}


\abstract{Parkinson’s disease (PD) is a chronic and complex neurodegenerative disorder influenced by genetic, clinical, and lifestyle factors. Predicting this disease early is challenging because it depends on traditional diagnostic methods that face issues of subjectivity, which commonly delay diagnosis. Several objective analyses are currently in practice to help overcome the challenges of subjectivity; however, a proper explanation of these analyses is still lacking. While machine learning (ML) has demonstrated potential in supporting PD diagnosis, existing approaches often rely on subjective reports only and lack interpretability for individualized risk estimation. This study proposed SCOPE-PD, an explainable AI-based prediction framework by integrating subjective and objective assessments to provide personalized  health decisions. 
Subjective and objective clinical assessment data are collected from the Parkinson’s Progression Markers Initiative (PPMI) study to construct a multimodal prediction framework. Several ML techniques were applied to these data, and the best ML model was selected to interpret the results. Model interpretability was examined using SHAP-based analysis.
The Random Forest algorithm achieved the highest accuracy of 98.66\% while working with the combined features from both subjective and objective test data. Tremor, bradykinesia, and facial expression are identified as the top three contributing features from the MDS-UPDRS test in the prediction of PD.}

\keywords{Parkinson's Disease, Machine Learning, Explainable Artificial Intelligence, Precision treatment, Clinical decision-making, Multi-modal Analysis.}



\maketitle

\section{Introduction}
 
Parkinson’s disease (PD) has emerged as one of the most pressing neurodegenerative challenges worldwide, affecting an estimated 11.8 million people as of 2021—a number projected to rise dramatically to over 25 million by 2050 as the global population ages \cite{luo2025global}. This rise not only puts a lot of stress on people and society, but it also makes the need for a quick and accurate diagnosis even more urgent, especially since there is still no cure that can improve the course of the disease. PD usually shows up as a mix of motor and non-motor symptoms caused by neurodegenerative changes, including the death of dopaminergic neurons and the buildup of alpha-synuclein aggregates~\cite{yaribash2025alpha}. The effects of PD on people and their families go beyond the emotional, cognitive, and physical effects. Additionally, the cost involved in diagnosing, monitoring, and treating PD is also rising, and it is now thought to be over USD 50 billion a year in the United States alone~\cite{chaudhuri2024economic}, which points to the need for earlier and more accurate diagnosis.

Even though neurological assessment has reached an advanced level, diagnostic accuracy is still low, at 70\%-80\%, especially in early stages. Clinical assessments are a big part of PD care and research, as they are the basis for diagnosis and the main way to record how the disease is getting worse. Traditionally, these assessments comprise a combination of subjective tests (e.g., patient-reported outcomes, questionnaires) and objective measures (e.g., clinician-rated motor and cognitive examinations). Although subjective and objective tools work well together to help doctors make decisions, their differences lead to ongoing debates about the best way to do things~\cite{siciliano2021correlates}. Finding a balance is important for both the accuracy of research and the care of patients. Recent evidence also suggests that patient-reported outcomes can meaningfully reflect underlying biological mechanisms, as demonstrated in REDONE-PD, where dopamine-related genetic mutations showed measurable associations with neurocognitive and functional outcomes in both PD patients and healthy controls~\cite{islam2024redone}.

Numerous studies have leveraged ML for the early detection of PD, harnessing everything from clinical diagnosis, sensor data, and voice characteristics to sophisticated neuroimaging. Grover et al. utilized voice recordings to accurately predict Parkinson's disease; however, medical history and cognitive data were excluded~\cite{grover2018predicting}. Templeton et al. found that perceived functionality by patients often differs from sensor-based metrics by applying ML while doing precise, stage-specific assessment and feature identification in PD~\cite{templeton2022classification}. Adebimp et al. developed an explainable ML pipeline for PD prediction using a multimodal dataset that combines demographics, medical history, lifestyle, symptoms, and standard clinical assessments (e.g., UPDRS, MoCA) with 93\% prediction accuracy~\cite{esan2025explainable}. Pereira et al. effectively employed movement analysis, yet they did not consider lifestyle and demographic factors~\cite{pereira2016new}.

A central barrier to broad clinical implementation of ML models lies in their “black box” nature. Without clear, patient-specific rationales for predictions, both clinicians and patients may be reluctant to trust or act on ML-driven insights. E\underline{X}plainable \underline{AI} (XAI) techniques, such as SHapley Additive exPlanations (SHAP)~\cite{lundberg2017unified}, address this issue by clarifying how predictions are made in complex health sectors (e.g., cancer \cite{sobhan2025tilda}, depression~\cite{tran2025explainable}). Adebimp et al., who applied SHAP for PD prediction, did not consider explaining from patient-reported and expert-assessed aspects.

\begin{figure*}[!h]
    \centering    \includegraphics[width=0.95\textwidth]{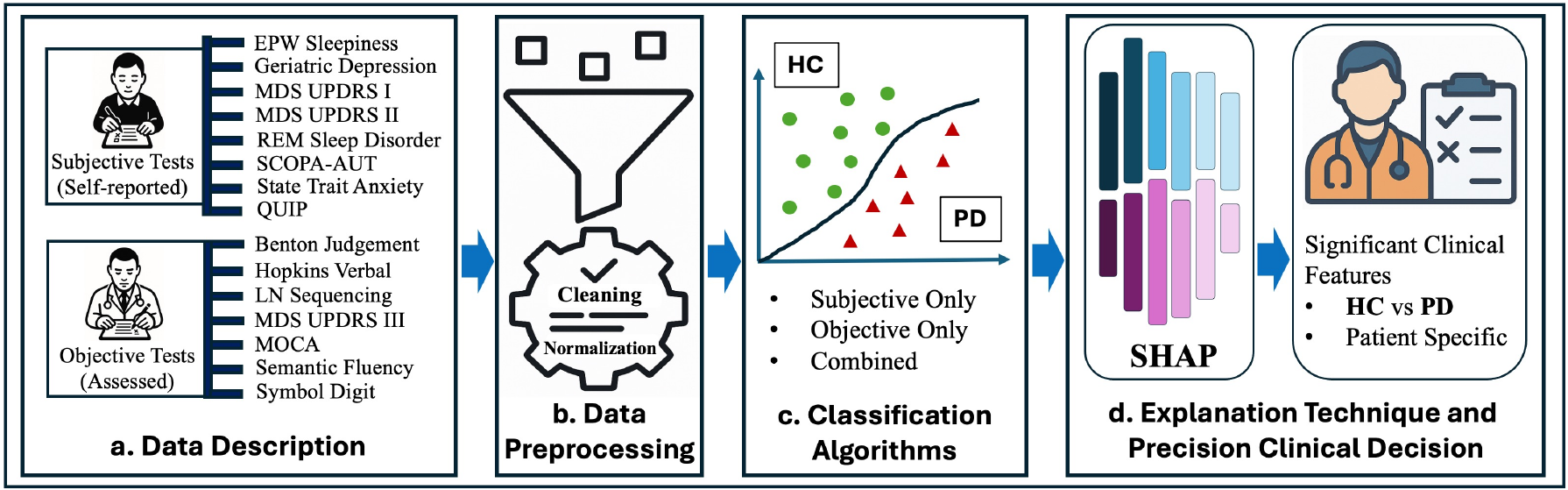}
    \caption{\textbf{Overall Study Framework including Data Description, Data Preprocessing, Classification Algorithms, and Explanation Technique and Precision Clinical Decision.} EPW: Epworth Sleepiness Scale; MDS-UPDRS: Movement Disorder Society Unified Parkinson's Disease Rating Scale; REM: Rapid Eye Movement; SCOPA-AUT: Scales for Outcomes in Parkinson’s disease - Autonomic Dysfunction; QUIP: Questionnaire for Impulsive-Compulsive Disorders; LN: Letter-Number; MOCA: Montreal Cognitive Assessment.
    }
    \label{fig:overall-study-framework}
\end{figure*}

In response to these challenges, this study seeks to reconcile the disparities between conventional clinical practice and emerging computational techniques by developing and validating an interpretable machine learning framework for Parkinson's disease diagnosis. This study aims to achieve two primary objectives: (1) to enhance the predictive accuracy of PD by optimally integrating diverse measures from the well-known PPMI repository, including both subjective self-reports and objective expert assessments; and (2) to offer clear, clinically relevant explanations for each model’s decision-making process, utilizing advanced SHAP tools.

The research illustrates, utilizing actual clinical data, that the integration of both data types enhances diagnostic accuracy. It identifies the most important features, like tremor (from subjective tests) and bradykinesia and facial expression (from objective tests), and also shows how they affect predictions. This approach not only makes ML more useful for diagnosing PD, but it also builds trust and makes it easier to understand, setting a standard for future explainable AI frameworks in neurodegenerative disease research.

\textbf{The key contributions of this study are:}
\begin{enumerate}
    \renewcommand{\labelenumi}{\alph{enumi}.}
    \item Assembled subjective and objective features from PPMI to train and compare state-of-the-art classifiers under nested cross-validation.

    \item Embedded SHAP to quantify both local (patient-specific) and global (cohort-level) feature contributions in probability space, enabling intuitive statements like “this feature raised the PD probability by +0.05.”

    \item Separated global feature contributions for Healthy Control (HC) and PD cohorts and by data type, making it clear whether the subjective report or the objective assessment influences the differences.

    \item Identified well-known motor determinants (e.g., tremor, bradykinesia) and quantified their relative influence alongside daily and autonomic functions

\end{enumerate}

\section{Dataset and Methodology}
The data used in this study were obtained from the Parkinson’s Progression Markers Initiative (PPMI) \cite{nalls2015diagnosis}, a large-scale, longitudinal, and publicly available dataset designed to identify biomarkers of PD progression. The study methodology can be found at ppmi-info.org. The data were obtained from the Laboratory of NeuroImaging repository (LONI). The dataset includes a comprehensive collection of subjective (patient-reported) and objective (expert-assessed) neuropsychological and clinical assessments.

The overall framework of our study is divided into four principal sections as shown in Figure~\ref{fig:overall-study-framework}: data source description, data preprocessing, classification algorithms, and explanation technique and precision clinical decision.

\subsection{Data Description}
In the PPMI dataset, each participant undergoes repeated evaluations over multiple visits. To ensure consistency and to avoid within-subject temporal dependencies, only the baseline (i.e., first-visit) data were used in this analysis. It also ensures the data collection alignment for the early prediction approach. The age range of participants is 26 to 86. The dataset encompasses a diverse range of subjective and objective assessments that collectively capture autonomic, behavioral, psychological, executive, memory, motor, sensory, sleep, and speech characteristics of individuals with PD and healthy controls. A detailed summary of the included tests, sample counts, number of survey questions, and feature distributions is provided in Table \ref{tab:data_summary}.

Subjective measurements include eight standardized self-reported questionnaires: the Epworth Sleepiness Scale (EPW) \cite{johns1991new}, Geriatric Depression Scale (GDS) \cite{greenberg2012geriatric}, Movement Disorder Society–Unified Parkinson’s Disease Rating Scale Parts I and II (MDS-UPDRS I–II) \cite{goetz2008movement}, REM Sleep Behavior Disorder Questionnaire \cite{schenck1986chronic}, Scales for Outcomes in Parkinson’s Disease–Autonomic Dysfunction (SCOPA-AUT) \cite{visser2004assessment}, State–Trait Anxiety Inventory (STAI) \cite{spielberger1983state}, and Questionnaire for Impulsive-Compulsive Disorders (QUIP) \cite{weintraub2012questionnaire}. Together, these subjective tests yielded 79 features from 2,128 participants across eight tests. 

Objective measurements consist of seven expert-assessed neuropsychological and motor tests: the Benton Judgment of Line Orientation (BJLO) \cite{benton1978visuospatial}, Hopkins Verbal Learning Test (HVLT) \cite{benedict1998hopkins}, Letter–Number Sequencing (LNS) \cite{wechsler2008wechsler}, MDS-UPDRS Part III \cite{goetz2008movement}, Montreal Cognitive Assessment (MoCA) \cite{hobson2015montreal}, Modified Semantic Fluency (MSF) \cite{zarino2014new}, and the Symbol Digit Modalities Test (SDMT) \cite{smith2013symbol}. These objective evaluations provided 69 features across 1,862 participants present in all seven tests. After combining the subjective and objective tests, 1,786 participants were found in all tests, yielding a total of 148 features (i.e., 79 subjective + 69 objective), as shown in Table \ref{tab:data_summary}. The conversion of survey questions and responses into features and feature values is described in Section 2.2.2. Descriptions of these features are provided in the supplementary material.

\begin{table*}[h]
\centering
\resizebox{0.95\linewidth}{!}{%
\begin{tabular}{@{}l*{9}{c}*{8}{c}@{}c}
\toprule
\textbf{Test}& \multicolumn{9}{c}{\textbf{Subjective Tests}} & \multicolumn{8}{c}{\textbf{Objective Tests}} & \textbf{Overall} \\
\cmidrule(lr){1-1}\cmidrule(lr){2-10} \cmidrule(l){11-18} \cmidrule(lr){19-19}
\textbf{Test Name} & 
\rotatebox{90}{EPW Sleepiness \cite{johns1991new}} & \rotatebox{90}{Geriatric Depression \cite{greenberg2012geriatric}}  & \rotatebox{90}{MDS UPDRS I \cite{goetz2008movement}}  & \rotatebox{90}{MDS UPDRS II \cite{goetz2008movement}}  & \rotatebox{90}{REM Sleep Disorder \cite{schenck1986chronic}}  & \rotatebox{90}{SCOPA-AUT \cite{visser2004assessment}}  & \rotatebox{90}{State Trait Anxiety\cite{spielberger1983state}}  & \rotatebox{90}{QUIP \cite{weintraub2012questionnaire}} & \rotatebox{90}{\textbf{\# of CSS / Total}}
 & 
\rotatebox{90}{Benton Judgement \cite{benton1978visuospatial}} & \rotatebox{90}{Hopkins Verbal \cite{benedict1998hopkins}} & \rotatebox{90}{LN Sequencing \cite{wechsler2008wechsler}} & \rotatebox{90}{MDS UPDRS III \cite{goetz2008movement} } & \rotatebox{90}{MOCA \cite{hobson2015montreal}} & \rotatebox{90}{Semantic Fluency \cite{zarino2014new}} & \rotatebox{90}{Symbol Digit \cite{smith2013symbol}}  & \rotatebox{90}{\textbf{\# of COS / Total}}
 & \rotatebox{90}{\textbf{\# of OCS / Total}} \\

\midrule
\textbf{\# of Samples} & 
3,832 & 3,761 & 2,212 & 2,212 & 3,839 & 3,830 & 3,760 & 3,829 & \textbf{2,128} & 3,813 & 3,819 & 3,813 & 3,817 & 3,813 & 2,562 & 2,370 & \textbf{1,862} & \textbf{1,786} \\
\textbf{\# of Survey Questions} & 
8 & 15 & 7 & 13 & 21 & 21 & 40 & 13 & \textbf{138} & 15 & 7 & 7 & 32 & 27 & 3 & 1 & \textbf{92} & \textbf{230} \\
\textbf{\# of Features} & 
1 & 1 & 7 & 13 & 21 & 21 & 2 & 13 & \textbf{79} & 1 & 4 & 1 & 32 & 27 & 3 & 1 & \textbf{69} & \textbf{148} \\

\bottomrule
\end{tabular}%
}
\caption{\textbf{Summary of Subjective and Objective Tests.} EPW: Epworth Sleepiness Scale; MDS-UPDRS: Movement Disorder Society Unified
Parkinson’s Disease Rating Scale; REM: Rapid Eye Movement; SCOPA-AUT: Scales for Outcomes in Parkinson’s disease - Autonomic
Dysfunction; QUIP: Questionnaire for Impulsive-Compulsive Disorders; LN: Letter-Number; MOCA: Montreal Cognitive
Assessment; CSS: Common Subjective Samples; COS: Common Objective Samples; OCS: Overall Common Samples;}
\label{tab:data_summary}
\end{table*}

\subsection{Data Preprocessing}
\subsubsection{Handling missing data}
\begin{figure}[!h]
    \centering    \includegraphics[width=0.90\columnwidth]{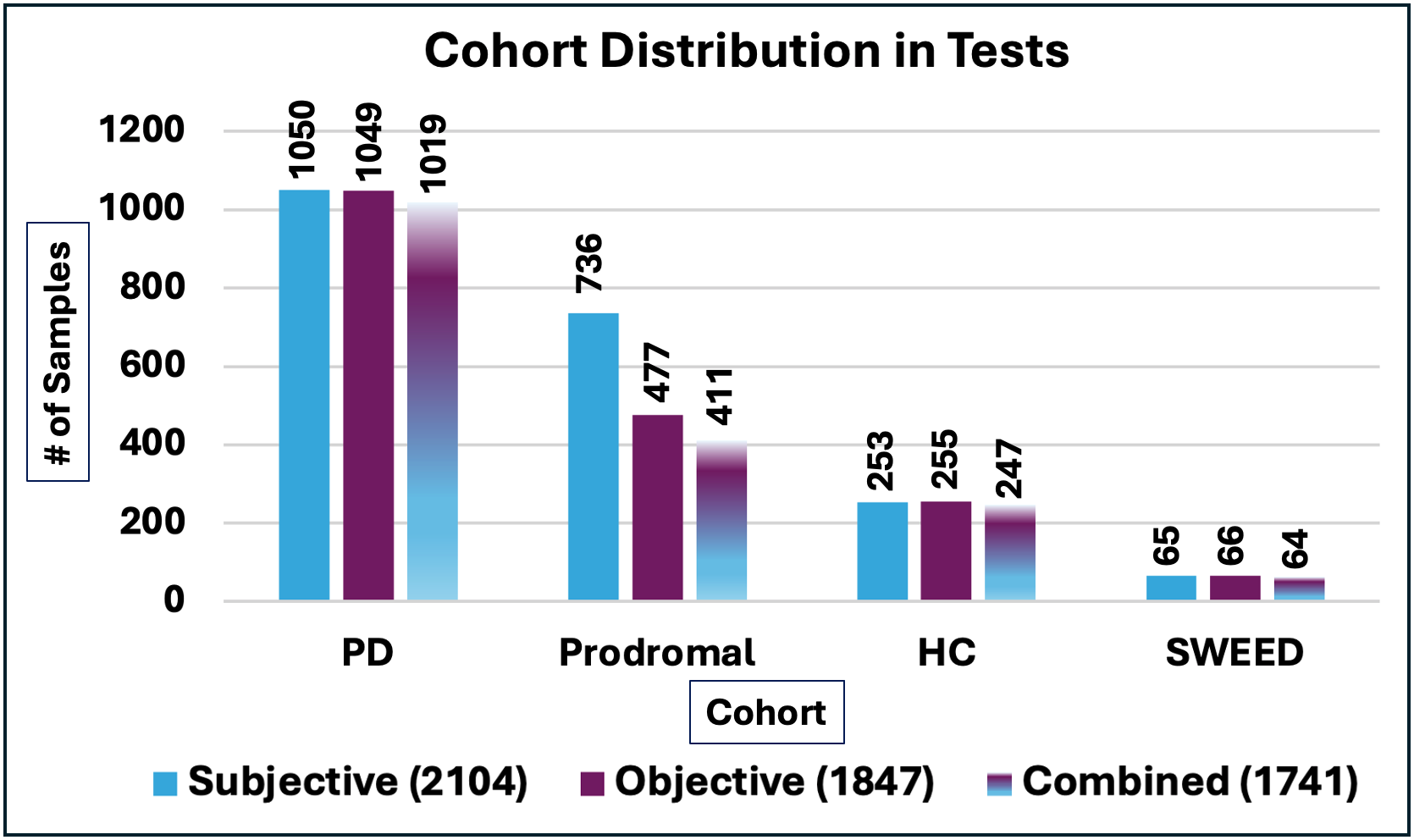}
    \caption{\textbf{Cohort description after data preprocessing.}}
    \label{fig:cohort}
\end{figure}

The preprocessing starts with removing samples and features with missing values. The ``VLTVEG'' (total correct vegetables an individual can list in 60 seconds) and ``VLTFRUIT'' (total correct fruits an individual can list in 60 seconds) features of the MSF test were found to be missing in 422 samples out of 1,862. For that reason, these two features were removed from the objective test features. After this step, the number of samples with subjective tests reduced from 2,128 to 2,104, and objective tests reduced from 1,862 to 1,847. The number of samples after combining subjective and objective tests is 1,741. Figure \ref{fig:cohort} summarizes the distribution of four cohorts (PD, Prodromal, HC, and SWEED) for three test types (Subjective, Objective, and Combined) in the cleaned data. For example, in subjective tests, 1,050 are PD patients (i.e., participants with a clinical diagnosis of PD), 736 are Prodromal (i.e., participants without a PD diagnosis yet but who show risk factors or early non-motor symptoms), 253 are age-matched healthy controls (HC) (i.e., participants with no known neurological symptoms or risk factors), and 65 are Scans Without Evidence of Dopaminergic Deficit (SWEDD). This study examined the PD and HC groups from three assessment conditions (subjective, objective, and combined) to identify the most significant features that contribute to precise clinical decisions.

\subsubsection{Converting survey responses into feature values}
The features considered in this study represent a certain question from a test or a combination of responses from multiple questions. In the case of subjective tests, the EPW test includes eight different questions related to sleep, which are converted to a single value by adding the response values. The GDS test has 15 questions for patients related to depression, which are converted to a single value by adding them. The 20 raw responses from the MDS-UPDRS I and II are included as features. Similarly, the raw responses to 13 behavior/psychology-related questions from the QUIP test are considered as features. The REM test asks 21 questions related to sleep quality and the presence of certain diseases in a patient. One of them asks about the presence of ``PARKISM'' in a patient, which may bias the study for ML-based analysis. We dropped this feature and moved on with 20 features from the REM test. The SCOPA-AUT test asks 21 questions related to autonomic function, which are considered as features directly. The STAI test includes 40 questions, where 21 questions follow the natural flow of response value, where a higher score relates to a higher anxiety level by asking questions related to anxiety. The remaining 19 questions are related to reverse situations where a higher score relates to a lower anxiety level. To align the response values, we reversed the value of 19 questions. Among the 40 questions, the first 20 questions are related to state (how a person feels ``right now'' or ``at this moment,'' reflecting temporary, situational anxiety), and the last 20 questions are related to trait (how a person generally feels, reflecting more stable, long-term anxiety tendencies). State and trait questions are converted to two single values by summing up the corresponding response values. So, in total, the subjective tests have 79 features to study.

Among the objective tests, the summation of 15 assessment scores related to executive function from the BJLO test is considered as a single feature. The HVLT test had seven assessment scores, ranging from 0 to 12, related to memory in the original assessment. To summarize the HVLT performance and reduce noise, they are converted to four different scores to consider as features. The first score established overall acquisition ability by summing up 3 individual recalls with a range of 0 to 36. The second one ranges from 0 to 12, reflecting the number of words recalled after a delay. The third one measures how much information in percentage is retained over delay. The final one scores the recognition discrimination index by subtracting false positives from true positives. The LNS test measures a patient's working memory and executive function with seven scores, which are converted to a single feature value by adding them. The MDS-UPDRS III test includes 32 questions related to motor function, which are directly considered as features. MOCA has 27 questions designed as a rapid screening instrument for mild cognitive dysfunction, which are directly used as features. MSF has three features, from which 2 of them are removed for missing values and only one is kept. The SDMT test assesses cognitive function by measuring how quickly a person can match symbols to numbers, and the total number of correct matches is used as a feature value. So, in total, the objective tests have 67 features to study. Here all the test response values except the MDS-UPDRS III follow the same pattern of higher scores representing better conditions of individual participants. To align all the objective and subjective tests, the scores are reversed so a higher score represents worse conditions.

\subsubsection{Normalizing feature values}
The response scores vary from test to test. For example, responses to MDS-UPDRS test questions are 0 to 4, and MOCA test questions are 0 to 1. To align the response values, all numeric features were standardized using min-max normalization to achieve uniformity within a 0-1 range. 

The final outcome of the data preprocessing step resulted in three datasets: subjective, objective, and combined. These three curated baseline dataset forms the foundation for classification and explainability analyses within the SCOPE-PD framework. 

\subsection{Classification Algorithms for PD Prediction}
To explore different machine learning strategies for differentiating PD from HC, five widely used supervised algorithms were implemented: Logistic Regression (LR), Support Vector Machine (SVM) with a radial-basis kernel, K-Nearest Neighbors (KNN), Random Forest (RF), and Extreme Gradient Boosting (XGBoost). All analyses were carried out in Python using the scikit-learn and XGBoost libraries. 

Since all three datasets (subjective, objective, and combined) are imbalanced, as shown in Figure \ref{fig:cohort}, native weighting was used instead of data augmentation using SMOTE and ADASYN to avoid synthetic sample bias. For LR, SVM, and RF, the argument \textbf{class\_weight=``balanced''} was applied to automatically adjust each class’s contribution to the loss function in proportion to its frequency. For XGBoost, imbalance was managed using the parameter \(\textbf{scale\_pos\_weight}\) as shown in Equation \ref{eq:XGB_imbalance}. This parameter scales the gradient and Hessian for the positive class during training, effectively increasing the contribution of the minority class to the loss function. When the dataset is highly imbalanced (i.e., \( n_0 \gg n_1 \)), setting a larger \(\text{scale\_pos\_weight}\) ensures that misclassified positive samples receive a higher penalty. As a result, the model becomes more sensitive to the minority class and can achieve better recall and balanced performance across both classes. KNN, which does not include a native weighting mechanism, was evaluated directly on the stratified data.
\begin{equation}
    \text{scale\_pos\_weight} = \frac{n_0}{n_1}
    \label{eq:XGB_imbalance}
\end{equation}
\textbf{where}
\begin{itemize}
\item \textbf{\( n_0 \)}: Number of negative samples
\item \textbf{\( n_1 \)}: Number of positive samples
\end{itemize}

Hyperparameter tuning and model evaluation followed a 5-fold stratified cross-validation framework. The dataset was first divided into an 80/20 train–test split, stratified to preserve class ratios. The training subset (80\%) was used exclusively for model development and hyperparameter optimization through a 5-fold cross-validation (GridSearchCV) procedure, which exhaustively searched predefined parameter grids for each algorithm using the F1-score as the optimization metric to balance sensitivity and precision. The best-performing hyperparameter configuration, identified by the highest mean cross-validated F1-score across the training folds, was then retrained on the entire 80\% training data and evaluated once on the held-out 20\% test set, which served as an independent set for unbiased performance assessment.





\subsection{Model Explainability using SHAP}
Local explanations were reported with SHAP waterfall/force plots for individual patients (showing how their feature values shift the prediction from the baseline to the final probability) and global explanations by aggregating attributions across the cohort. Specifically, for each feature, mean absolute SHAP was computed for the HC and PD groups separately, and these were visualized as stacked bars (HC + PD). This class-conditional aggregation clarifies not only which features are most important overall, but also in which group their importance primarily arises. The concept of local interpretability of SHAP was utilized for precision clinical decision-making that may assist in formulating strategies for individualized treatment.

Overall, the use of SHAP enabled a comprehensive understanding of both global feature importance and individual-level predictions. By revealing the roles of subjective and objective features in model decision-making, this explainable AI framework strengthened the interpretability, clinical relevance, and trustworthiness of the proposed PD prediction system.



\section{Results and Discussion}
The results are divided into two sections: a. experimental results of ML models and b. interpretation of ML predictions for Parkinson's disease based on subjective, objective, and combined features. 
\subsection{Experimental Results of Machine Learning Models}
Model performance varied across feature categories, revealing distinct predictive patterns between subjective, objective, and combined feature representations. The performance evaluation values are listed in Table~\ref{tab:model_eval}. For subjective assessments, the RF model achieved the highest overall performance with an F1-score of 0.9786 ± 0.0016, ROC-AUC of 0.9875 ± 0.0032, and PR-AUC of 0.9970 ± 0.0008, indicating strong discriminative power and robustness in capturing non-linear feature interactions from patient-reported measures. The SVM (RBF) and XGBoost models followed closely, with comparable precision and recall, while LR and KNN exhibited slightly lower but still competitive results, confirming that both linear and non-linear classifiers can effectively separate PD and healthy control groups when subjective scales are well standardized.

When using objective cognitive and motor features, performance improved further across most algorithms. RF again outperformed others with a mean accuracy of 0.9785 ± 0.0069 and recall of 0.9924 ± 0.0072, demonstrating excellent sensitivity to PD-related performance impairments. SVM and XGBoost achieved similar AUC values above 0.99, suggesting consistent generalizability and reliability across folds. These findings highlight that objective test metrics offer slightly superior predictive value compared to subjective measures.

When subjective and objective features were combined, model performance reached its peak across all metrics. The RF had the best overall accuracy (0.9866 ± 0.0091) and F1-score (0.9917 ± 0.0055). The ROC-AUC (0.9992 ± 0.0006) and PR-AUC (0.9998 ± 0.0001) showed that it could almost perfectly tell the difference between the two groups. The SVM (RBF) and XGBoost followed closely, maintaining mean F1-scores above 0.988, while LR also performed strongly (F1 = 0.9871 ± 0.0045), confirming the complementary predictive value of integrating subjective and objective features. Although KNN achieved the highest precision (0.9978 ± 0.0050), its comparatively lower recall (0.8371 ± 0.0217) suggests limited sensitivity to PD cases despite its specificity toward healthy controls. 

\begin{figure}[!htbp]
    \centering    \includegraphics[width=0.95\linewidth]{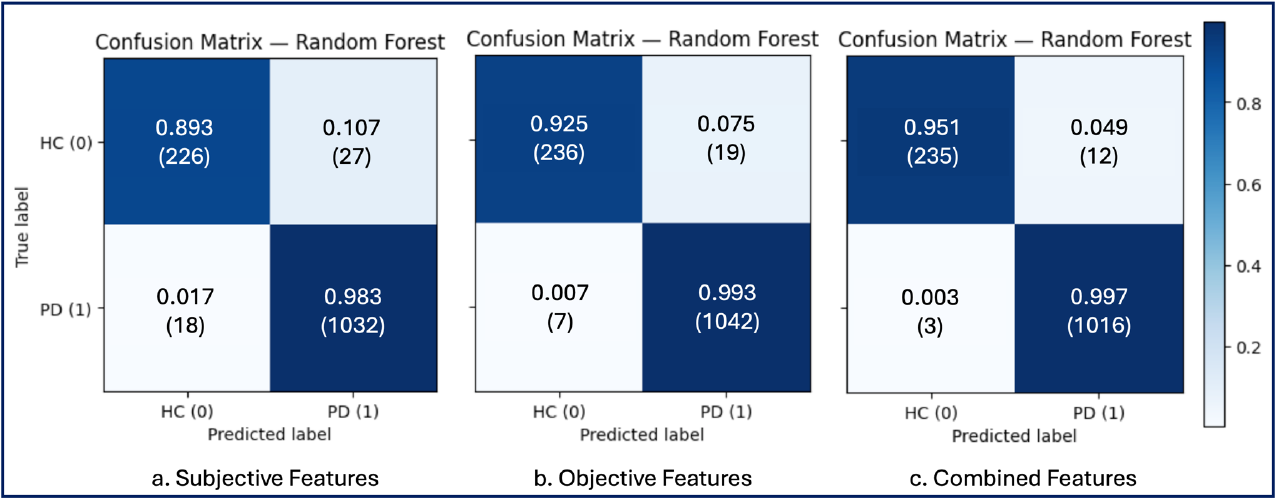}
    \caption{\textbf{Out-of-Fold (OOF) confusion matrix (normalized) for the best ML model.} RF achieved the highest accuracy with all three types of features—subjective, objective, and combined.
    }
    \label{fig:confusion_matrix}
\end{figure}

Although the reported accuracy is high, it was obtained under strict 5-fold nested cross-validation using class-level stratification, ensuring no overfitting to data noise. Thus, the high model performance reflects that the survey responses at the first visit can clearly distinguish PD from HC groups. Nevertheless, we recognize that external validation on independent cohorts is essential to confirm real-world performance. Overall, across all experiments, ensemble models such as RF and XGBoost consistently demonstrated superior and more stable performance than single estimators, emphasizing the importance of non-linear modeling and feature interaction learning in PD classification tasks. To further validate classification reliability and generalization, confusion matrices were generated for the best-performing models (Random Forest in all three datasets) using the Out-of-Fold (OOF) strategy (Figure \ref{fig:confusion_matrix}). The number of misclassified samples decreased from the models with the subjective features to the objective features and improved further with the combined features.

\begin{table*}[h]
\centering
\caption{Performance comparison across classification models for Parkinson’s disease prediction using Subjective, Objective, and Combined features (mean $\pm$ standard deviation over 5 folds). SD: Standard Deviation.}
\label{tab:model_eval}
\resizebox{\linewidth}{!}{
\begin{tabular}{llcccccc}
\toprule
\textbf{Test Type} & \textbf{Model} & \textbf{Accuracy $\pm$ SD} & \textbf{Precision $\pm$ SD} & \textbf{Recall $\pm$ SD} & \textbf{F1 Score $\pm$ SD} & \textbf{ROC AUC $\pm$ SD} & \textbf{PR AUC $\pm$ SD} \\
\midrule
\multirow{5}{*}{\textbf{Subjective}} 
 & Logistic Regression & 0.9440 $\pm$ 0.0111 & 0.9851 $\pm$ 0.0034 & 0.9448 $\pm$ 0.0141 & 0.9645 $\pm$ 0.0072 & 0.9820 $\pm$ 0.0047 & 0.9956 $\pm$ 0.0013 \\
 & KNN & 0.9010 $\pm$ 0.0150 & 0.9610 $\pm$ 0.0106 & 0.9143 $\pm$ 0.0121 & 0.9370 $\pm$ 0.0096 & 0.9525 $\pm$ 0.0148 & 0.9881 $\pm$ 0.0041 \\
 & SVM (RBF) & 0.9563 $\pm$ 0.0043 & 0.9779 $\pm$ 0.0052 & 0.9676 $\pm$ 0.0098 & 0.9727 $\pm$ 0.0029 & 0.9778 $\pm$ 0.0123 & 0.9936 $\pm$ 0.0049 \\
 & Random Forest & \textbf{0.9655 $\pm$ 0.0027} & 0.9755 $\pm$ 0.0074 & \textbf{0.9819 $\pm$ 0.0062} & \textbf{0.9786 $\pm$ 0.0016} & \textbf{0.9875 $\pm$ 0.0032} & \textbf{0.9970 $\pm$ 0.0008} \\
 & XGBoost & 0.9609 $\pm$ 0.0079 & \textbf{0.9846 $\pm$ 0.0101} & 0.9667 $\pm$ 0.0101 & 0.9755 $\pm$ 0.0049 & 0.9831 $\pm$ 0.0037 & 0.9954 $\pm$ 0.0016 \\
\midrule
\multirow{5}{*}{\textbf{Objective}} 
 & Logistic Regression & 0.9663 $\pm$ 0.0099 & 0.9883 $\pm$ 0.0055 & 0.9695 $\pm$ 0.0079 & 0.9788 $\pm$ 0.0062 & 0.9928 $\pm$ 0.0013 & 0.9982 $\pm$ 0.0002 \\
 & KNN & 0.9026 $\pm$ 0.0155 & \textbf{0.9905 $\pm$ 0.0057} & 0.8875 $\pm$ 0.0213 & 0.9360 $\pm$ 0.0109 & 0.9618 $\pm$ 0.0144 & 0.9859 $\pm$ 0.0061 \\
 & SVM (RBF) & 0.9778 $\pm$ 0.0074 & 0.9857 $\pm$ 0.0058 & 0.9867 $\pm$ 0.0040 & 0.9862 $\pm$ 0.0046 & 0.9928 $\pm$ 0.0021 & 0.9980 $\pm$ 0.0008 \\
 & Random Forest & \textbf{0.9785 $\pm$ 0.0069} & 0.9812 $\pm$ 0.0085 & \textbf{0.9924 $\pm$ 0.0072} & \textbf{0.9867 $\pm$ 0.0043} & \textbf{0.9969 $\pm$ 0.0019} & \textbf{0.9991 $\pm$ 0.0007} \\
 & XGBoost & 0.9739 $\pm$ 0.0057 & 0.9858 $\pm$ 0.0109 & 0.9819 $\pm$ 0.0085 & 0.9838 $\pm$ 0.0035 & 0.9955 $\pm$ 0.0018 & 0.9988 $\pm$ 0.0006 \\
\midrule
\multirow{5}{*}{\textbf{Combined}} 
 & Logistic Regression & 0.9795 $\pm$ 0.0071 & 0.9931 $\pm$ 0.0055 & 0.9813 $\pm$ 0.0107 & 0.9871 $\pm$ 0.0045 & 0.9972 $\pm$ 0.0021 & 0.9993 $\pm$ 0.0005 \\
 & KNN & 0.8673 $\pm$ 0.0150 & \textbf{0.9978 $\pm$ 0.0050} & 0.8371 $\pm$ 0.0217 & 0.9102 $\pm$ 0.0112 & 0.9744 $\pm$ 0.0059 & 0.9907 $\pm$ 0.0021 \\
 & SVM (RBF) & 0.9818 $\pm$ 0.0045 & 0.9912 $\pm$ 0.0085 & 0.9863 $\pm$ 0.0094 & 0.9887 $\pm$ 0.0028 & 0.9984 $\pm$ 0.0006 & 0.9996 $\pm$ 0.0001 \\
 & Random Forest & \textbf{0.9866 $\pm$ 0.0091} & 0.9875 $\pm$ 0.0137 & \textbf{0.9961 $\pm$ 0.0041 }& \textbf{0.9917 $\pm$ 0.0055} & \textbf{0.9992 $\pm$ 0.0006} & \textbf{0.9998 $\pm$ 0.0001} \\
 & XGBoost & 0.9826 $\pm$ 0.0045 & 0.9932 $\pm$ 0.0073 & 0.9853 $\pm$ 0.0092 & 0.9892 $\pm$ 0.0028 & 0.9988 $\pm$ 0.0006 & 0.9997 $\pm$ 0.0001 \\
\bottomrule
\end{tabular}
}
\end{table*}

\subsection{Interpretation of ML Predictions for PD}
To provide transparent, clinically meaningful explanations of the tuned Random Forest, we used SHAP TreeExplainer, a tree-model–aware method that computes exact, locally additive feature attributions for decision-tree ensembles~\cite{lundberg2020local}. TreeExplainer provides, for each subject, an attribution that indicates how much each feature increases or decreases the model’s predicted probability of PD compared to a reference expectation (“baseline”). For binary classification, TreeExplainer returns one attribution vector per class. The PD channel (class = 1) was analyzed, where positive values push the prediction toward PD and negative values toward HC (class = 0).

\begin{figure*}[h]
    \centering    \includegraphics[width=\linewidth]{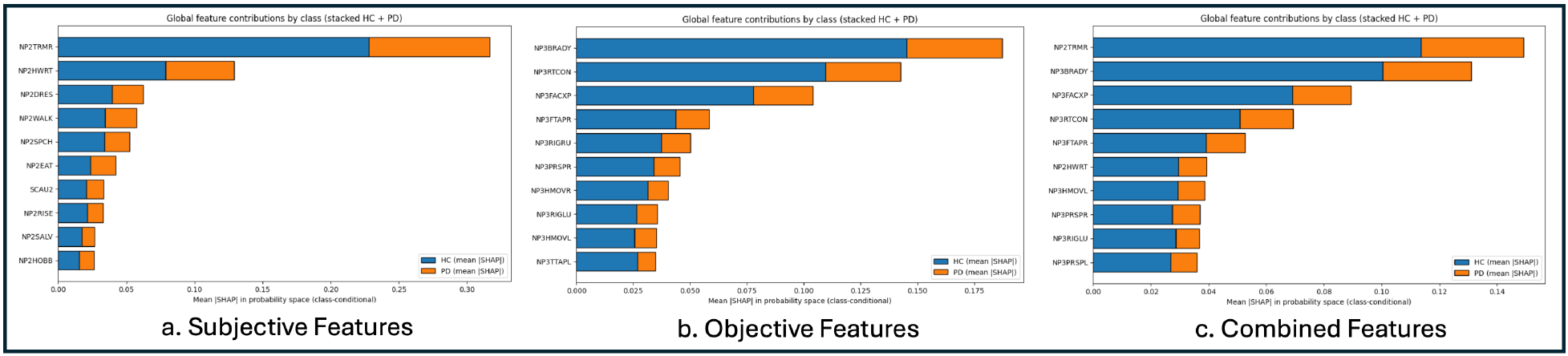}
    \caption{\textbf{Global feature contribution:} Top 10 features contributing towards HC and PD cohorts. a. Subjective features. b. Objective features. c. Combined features.
    }
    \label{fig:gfc_top_10}
\end{figure*}

Inputs to SHAP were the scaled features from the pipeline so that the attributions reflect exactly what the classifier sees. The contribution of features was analyzed in two aspects: global (contribution in predicting HC or PD) and local (contribution in a sample). The stacked bar plot (Figure \ref{fig:gfc_top_10}) shows the top 10 feature contributions towards class-conditional decisions for the features of subjective, objective, and combined tests. In the subjective dataset, 9 out of 10 top contributing features are from MDS-UPDRS II and one is from SCOPA-AUT (Figure \ref{fig:gfc_top_10}a). It is clear from this figure that the self-reported tremor feature (NP2TRMR) dominated the importance ranking by a wide margin within both HC and PD groups. The next tier comprised handwriting difficulty (NP2HWRT), dressing (NP2DRES), walking/balance (NP2WALK), speech (NP2SPCH), and eating/feeding (NP2EAT). These day-to-day activities showed mixed HC/PD splits: very low difficulty supports HC, whereas increasing difficulty contributes evidence toward PD. The SCOPA-AUT feature SCAU2 (In the past month, has saliva dribbled out of your mouth?) was found to be one of the most contributing features. 

The top 10 features for objective tests are coming from MDS-UPDRS III test (Figure \ref{fig:gfc_top_10}b), where the top contributed feature (NP3BRADY) related to observing ``spontaneous gestures while sitting, and the nature of arising and walking,'' which is a significant PD sign and central to diagnosis/staging. So, its dominance as a driver of PD is expected and provides the significance of this study. It also aligns with next-ranked items: NP3RTCON (rapid alternating movement/coordination), NP3FACXP (facial expression), and NP3FTAPR (finger tapping), which are closely related to bradykinetic/akinetic measures. While combining subjective and objective tests, eight features from MDS-UPDRS III and two features from MDS-UPDRS II appeared in the top 10 contributed features, as shown in \ref{fig:gfc_top_10}c. The self-reported tremor impact (NP2TRMR) was found to be the most contributing in distinguishing PD and HC samples.

\begin{figure}[h]
    \centering    \includegraphics[width=0.95\linewidth]{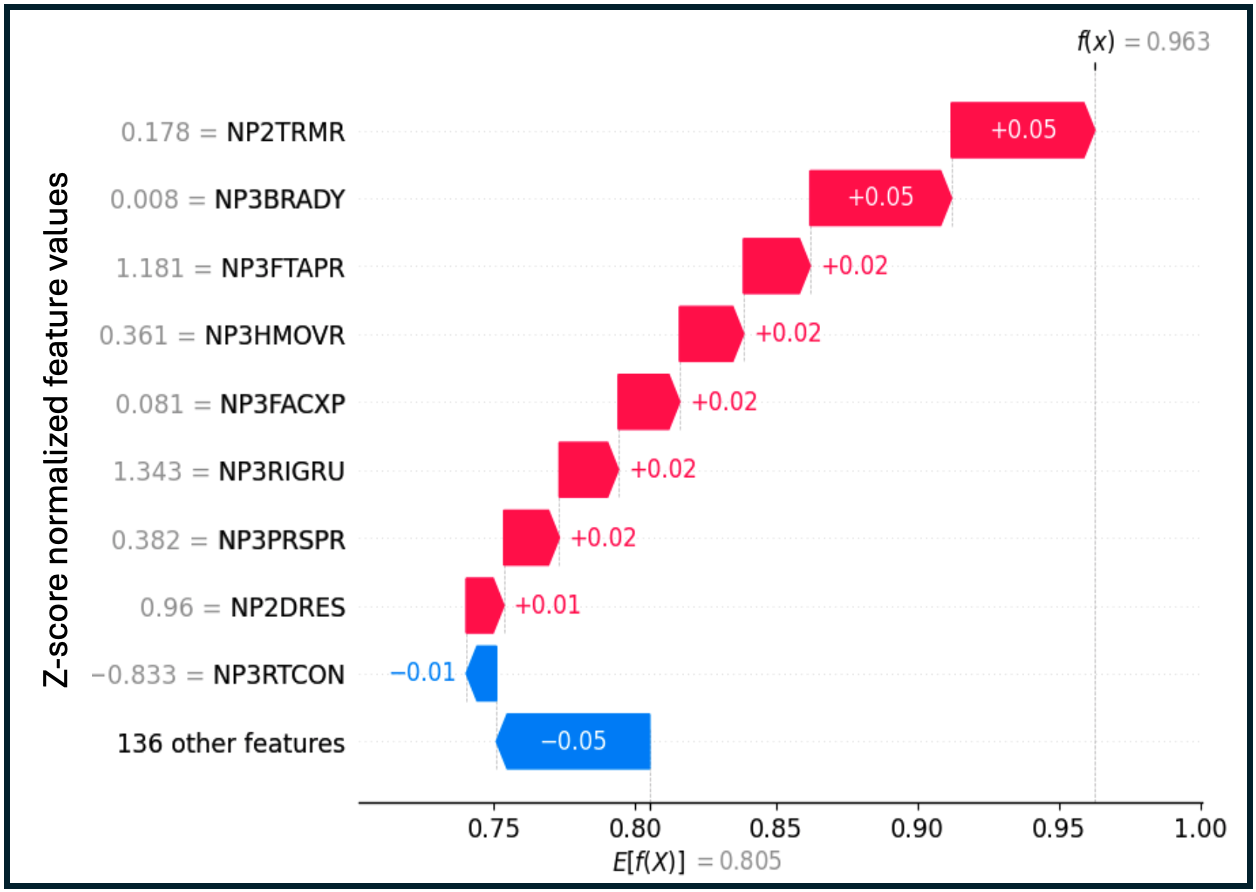}
    \caption{\textbf{Local feature contribution:} Top 10 local features contributing to a specific PD patient. 
    }
    \label{fig:local_combined}
\end{figure}

To gain a deeper understanding of the model's decision-making process, we examined specific individuals diagnosed as PD patient, as shown in Figure \ref{fig:local_combined}. We utilized SHAP's color-coded visualizations to identify the features that significantly influenced the model's prediction for each person. Red hues highlight features that strongly contribute in diagnosing PD, while
blue hues indicate features aligned with HC prediction. The model’s baseline probability of PD was 0.805 and rose to 0.963 once the patient’s features were considered. The increase was driven mainly by tremor (NP2TRMR, +0.05) and bradykinesia–related (NP3BRADY, +0.05) markers. Other positive features are finger tapping (NP3FTAPR, +0.02), hand movements (NP3HMOVR, +0.02), masked facies/facial expression (NP3FACXP, +0.02), upper-limb rigidity (NP3RIGRU, +0.02), posture (NP3PRSPR, +0.02), and dressing difficulty (NP2DRES, +0.01). Rapid alternating movements/coordination (NP3RTCON) countered the prediction ($-$0.05) meaning the PD patient had a slight issue with this activity. The remaining features had negligible effects. While SHAP predominantly highlighted established PD determinants such as tremor and bradykinesia, the framework also quantified their relative effect sizes across subjective and objective domains, offering clinicians quantitative thresholds for individualized interpretation. This transparency transforms well-known symptoms into measurable, patient-specific risk weights usable in clinical discussions.

\subsection{Discussion}
This study aimed to identify an effective model for predicting PD using self-reported and assessment-based tests and explain the contribution of features using SHAP based on the most effective model. Five individual models were evaluated: LR, SVM-RBF, KNN, RF, and XGBoost. RF attained the highest accuracy rates of 96.55\%, 97.85\%, and 98.66\% for subjective, objective, and combined tests, respectively. It positioned this study competitively within the existing literature to compare with other approaches as shown in Table \ref{tab:literature_compare}. Grover et al. reported 81.67\% accuracy by using only the speech features \cite{grover2018predicting}, while Templeton et al. achieved 92.6\% accuracy by analyzing sensor-based metric~\cite{templeton2022classification}. Adebimpe et al. \cite{esan2025explainable} achieved 93\% accuracy with clinical (MoCA), cognitive (UPDRS), demographic, and lifestyle data.  Dentamaro et al. \cite{dentamaro2024enhancing} attained an accuracy of 96.6\% by employing DenseNet on the detection of the prodromal stage by utilizing 3D MRI images and clinical data. Priyadharshini et al. \cite{priyadharshini2024comprehensive} established a complete framework utilizing T2-weighted 3D MRI datasets and attained an accuracy of 96.8\% through the integration of Gradient Boosting and SMOTE for data balancing. This study compares its findings with the existing literature by using multimodal self-reported and clinically assessed data from 1786 samples. To address the sample imbalance, this study utilized the default class-weight property for LR, SVM, and RF models, or the scale\_pos\_weight parameter for the XGBoost model, rather than applying SMOTE to generate synthetic data.

\begin{table*}[ht]
\centering
\caption{Comparison of the proposed SCOPE-PD model with existing studies on Parkinson's disease prediction.}
\label{tab:literature_compare}
\resizebox{\columnwidth}{!}{
\begin{tabular}{@{}l p{3.5cm} p{1.5cm} p{2.2cm} p{3cm}@{}}
\toprule
\textbf{Author} & \textbf{Data} & \textbf{Accuracy} & \textbf{Dataset} & \textbf{Classification} \\
\cmidrule(lr){1-1} \cmidrule(lr){2-5}

Grover et al.~\cite{grover2018predicting} & Voice recordings & 81.67\% & UCI ML Repository & Severe vs Not Severe \\
Templeton et al.~\cite{templeton2022classification} & Sensor-based metric & 92.60\% & Self-Collected & PD vs HC \\
Adebimpe et al.~\cite{esan2025explainable} & Subjective (UPDRS, MoCA) & 93\% & PPMI & PD vs HC \\
Dentamaro et al.~\cite{dentamaro2024enhancing} & 3D MRI Image, Clinical & 96.60\% & PPMI & Prodromal stages \\
Priyadharshini et al.~\cite{priyadharshini2024comprehensive} & T2-weighted 3D MRI & 96.80\% & PPMI & PD vs Prodromal vs HC \\

\midrule
\textbf{SCOPE-PD Subjective} & Subjective measures & 96.50\% & PPMI & PD vs HC \\

\textbf{SCOPE-PD Objective} & Objective measures & 97.80\% & PPMI & PD vs HC \\

\textbf{SCOPE-PD Combined} & Subjective + Objective & \textbf{98.60\%} & PPMI & PD vs HC \\
\bottomrule
\end{tabular}
}
\footnotesize
\end{table*}

\textbf{Limitations and Future Scopes}: Only baseline data (first visit) were used to ensure the existence of multimodal data, and progression modeling was beyond the current scope. Incorporating longitudinal trajectories from follow-up visits is a next step toward progression and prognosis prediction. An expanded version of SCOPE-PD will integrate genetic (e.g., GBA, SNCA) and MRI/fMRI imaging modalities from complementary cohorts to enable a full multimodal diagnostic framework. The absence of external validation datasets limits the ability to fully assess robustness and real-world applicability. Future efforts will incorporate multi-site data from PDBP and BioFIND for external validation. Deploying the model in clinical trials will facilitate the evaluation of its effects on diagnostic accuracy and patient care outcomes in practical environments.

\section{Conclusion}
The study SCOPE-PD aimed to develop and evaluate predictive models for PD applying ML techniques, corroborated by the explainable AI (XAI) method for better interpretability. Rather than claiming clinical deployment readiness, SCOPE-PD should be viewed as a methodological bridge linking subjective and objective markers through explainable modeling. Through a comprehensive evaluation of five ML models, this study found Random Forest with the highest performance, achieving accuracy above 96\% across all three datasets. The novelty lies in unifying both modalities under a single interpretable framework, not in outperforming prior work. The results should therefore be interpreted cautiously until externally validated.  The model could serve as a valuable screening tool for early PD detection, potentially identifying at-risk patients before significant symptom onset, as all data used here are collected at their first visit for diagnosis. The XAI analysis revealed that the MDS-UPDRS test features significantly contribute to the prediction of Parkinson's disease. The RF model's interpretability makes it appropriate for incorporation into clinical workflows, offering evidence-based diagnostic assistance. Moreover, the ability to identify individualized risk contributors opens opportunities for more personalized treatment planning and targeted interventions.

\textbf{Data Acknowledgment:} Data used in the preparation of this article were obtained [on May 3, 2025] from the Parkinson’s Progression Markers Initiative (PPMI) database (https://www.ppmi-info.org/access-data-specimens/download-data), RRID:SCR\_006431. 
For up-to-date information on the study, visit http://www.ppmi-info.org. PPMI – a public-private partnership – is funded by the Michael J. Fox Foundation for Parkinson’s Research and funding partners, including 4D Pharma, Abbvie, et al..

\textbf{Funding:} This work was partially supported by the NIH/NCI R21CA290324 and NIH/NHGRI UG3HG013615. The content is solely the responsibility of the authors and does not necessarily represent the official views of the funding agencies.

\textbf{Conflict of Interest Statement:}
All authors of this work declare that there are no conflicts of interest in the authorship nor publication of this contribution.

\bibliography{sn-bibliography}

\end{document}